\documentclass[sigconf,nonacm]{acmart}
\pdfoutput=1
\usepackage{epsfig}
\usepackage{graphicx}
\usepackage{subfigure}

\usepackage{amsmath}

\usepackage{bm}
\renewcommand\footnotecopyrightpermission[1]{}
\AtBeginDocument{%
  \providecommand\BibTeX{{%
    \normalfont B\kern-0.5em{\scshape i\kern-0.25em b}\kern-0.8em\TeX}}}

\begin{document}

\title{Adaptive Affinity Loss and Erroneous Pseudo-Label Refinement for Weakly Supervised Semantic Segmentation }



%
%
\author{Xiangrong Zhang}
\authornote{Corresponding author.}
\email{xrzhang@mail.xidian.edu.cn}
\affiliation{%
	\institution{School of Artificial Intelligence, Xidian University}
	\city{Xi’an}
	\country{China}
}

\author{Zelin Peng}
\email{zlpeng@stu.xidian.edu.cn}
\affiliation{%
	\institution{School of Artificial Intelligence, Xidian University}	
	\city{Xi’an}
	\country{China}
}

\author{Peng Zhu}
\author{Tianyang Zhang}
\email{zhupeng@stu.xidian.edu.cn}
\email{tianyangzhang@stu.xidian.edu.cn}
\affiliation{%
	\institution{School of Artificial Intelligence, Xidian University}	
	\city{Xi’an}
	\country{China}
}

\author{Chen Li}
\email{cli@xjtu.edu.cn}
\affiliation{%
	\institution{School of Computer Science and Technology, Xi’an Jiaotong University}	
	\city{Xi’an}
	\country{China}
}

\author{Huiyu Zhou}
\email{hz143@leicester.ac.uk}
\affiliation{%
	\institution{School of Informatics University of Leicester}  
	\city{Leicester}
	\country{UK}
}

\author{Licheng Jiao}
\email{lchjiao@mail.xidian.edu.cn}
\affiliation{%
	\institution{School of Artificial Intelligence, Xidian University}	
	\city{Xi’an}
	\country{China}
}

\renewcommand{\shortauthors}{Zhang and Peng, et al.}
\fancyhead{}

%
%
%
%



\begin{abstract}
Semantic segmentation has been continuously investigated in the last ten years, and majority of the established technologies are based on supervised models. In recent years, image-level weakly supervised semantic segmentation (WSSS), including single- and multi-stage process, has attracted large attention due to data labeling efficiency. In this paper, we propose to embed affinity learning of multi-stage approaches in a single-stage model. To be specific, we introduce an adaptive affinity loss to thoroughly learn the local pairwise affinity. As such, a deep neural network is used to deliver comprehensive semantic information in the training phase, whilst improving the performance of the final prediction module. On the other hand, considering the existence of errors in the pseudo labels, we propose a novel label reassign loss to mitigate over-fitting. Extensive experiments are conducted on the PASCAL VOC 2012 dataset to evaluate the effectiveness of our proposed approach that outperforms other standard single-stage methods and achieves comparable performance against several multi-stage methods.
\end{abstract}

\maketitle
\section{Introduction}
Semantic segmentation aims to predict pixel-wise classification results on images, which is one of the vital tasks in computer vision and machine learning. With the development of deep learning, a variety of Convolutional Neural Network (CNN) based semantic segmentation methods \cite{chen2018encoder,chen2017deeplab,long2015fully} have achieved promising successes. However, these methods need to collect pixel-level labels, which are time consuming, laborious, and computationally expensive. To alleviate pixel-level annotations, many works have been devoted to weakly supervised semantic segmentation (WSSS) which applies weak supervisions by, e.g., image-level labels \cite{paper1,paperMIL,pathak2015,paperijcv,paperstc,paperfickle,paper38,paper19,paper29}, scribbles \cite{paper24,paper36,papernorm}, and bounding boxes \cite{paper7,paper34,khoreva2017simple}. In this paper, we focus on the most challenging case under weakly-supervised settings where image-level labels are utilized for training WSSS models.

Existing algorithms mainly consist of two ways to perform on image-level WSSS, by improving the performance of the initial Class Activation Map (CAM) \cite{papercam}, or refining the initial response with an additional model. Since CAM models lead to incomplete initial masks, some recent methods \cite{paper3,paper2,paper5,paper100} aim to obtain more precise and broader activation regions to deal with this problem. Different from that, early methods \cite{paper37,paper40,paper29} follow the seed-and-expand principle to refine the initial seeds (e.g. CAMs), while recent methods adopt an affinity learning method \cite{paper1} for further enhancement.

\begin{figure}[t]
	
	\centering
	
	\includegraphics[width=8.091cm, height=3.906cm]{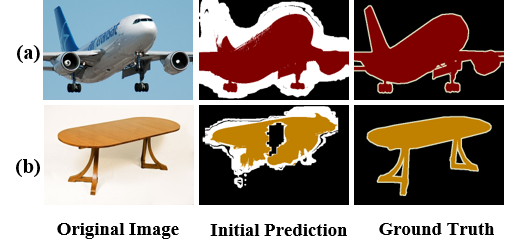}\\
	\caption{Conceptual illustration of initial predictions. Different colors represent confident areas of object classes and background: \textbf{red} for airplane, \textbf{orange} for table, and \textbf{black} for background. The neutral region is color-coded in \textbf{white}. \textbf{(a)} Ideal Initial Prediction, \textbf{(b)} Semantic Ambiguity. The general refining methods often have better results on \textbf{(a)} while several methods misfitted marks in \textbf{(b)}.}
	\label{fig. showmaker}
\end{figure}

Compared with the above methods, single-stage methods \cite{paperSSSS,paper4} are dedicated to efficient training in a straightforward manner. In spite of substantial segmentation accuracy and  streamlined process, these methods still fall behind most of the existing multi-stage models.
In fact, the quality of the initial labels closely determines the final segmentation results. Figure \ref{fig. showmaker} depicts certain diversity in initial labels: (\emph{i}) Figure \ref{fig. showmaker}(a), there are positive initial results compared with the ground truth. (\emph{ii}) Figure \ref{fig. showmaker}(b), mortal erroneous pixels exist. 
Most multi-stage methods severely rely on the fixed initial predictions that may limit their segmentation performance. As opposed to this, single-stage methods obtain initial seeds during training.


In this work, we leverage the merits of multi-stage methods to improve the performance of single-stage image-level semantic segmentation. Our intention is to refine the initial labels end-to-end via \emph{pixel-level affinity learning} and \emph{erroneous pixel revision} to improve the segmentation results. To be precise, we first create the Adaptive Affinity (AA) loss to enhance the process of semantic propagation. Instead of using a fixed confidence factor (i.e., 1), we here adopt adaptive weights in pairwise connections, which further suppresses the excessive attention on `fake' pixels. Second, we propose a novel Label Reassign (LR) loss that acts on the semantic propagation in the embedding space. This empowers the network to discern errors in initial labels, and thereby avoids over-fitting to the erroneous seeds.

Our proposed approach, stemming from the standard single-stage WSSS method \cite{paper4}, learns to improve the localization inference in a unified style. The main contributions of this work are summarized as follows:

$\bullet$ We present the AA loss to enhance the process of semantic propagation by computing connectivity while adaptively learning the local pairwise affinity.

$\bullet$ We propose the LR loss to identify erroneous initial predictions, which alleviates over-fitting on erroneous labels.

$\bullet$ Experiments on PASCAL VOC 2012 illustrate that our method achieves state-of-the-art performance in single-stage image-level WSSS.

\begin{figure*}[t]
	\centering
	\includegraphics[width=16.2364cm, height=7.2165cm]{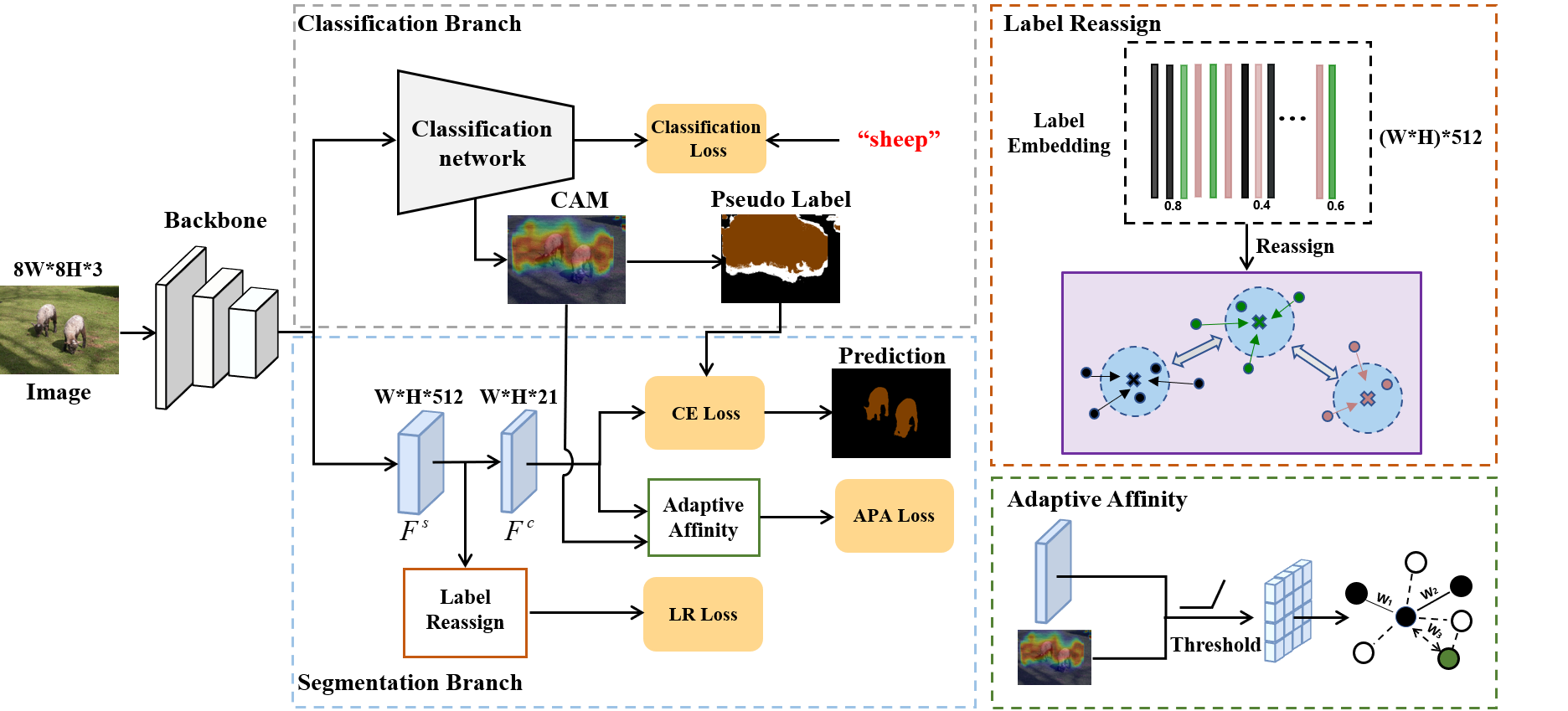}\\
	\caption{The proposed framework: Given an image, first of all, we create pseudo masks through the classification branch and then use the masks as segmentation supervision. Afterwards, we reassign the labeled point embedding and also derive the LR loss. Furthermore, we train the segmentation prediction with a novel adaptive affinity learning method and traditional cross entropy loss. The entire framework is optimized end-to-end during training.}
	\label{fig. geng}
\end{figure*}

\section{Related work}

\subsection{Weakly-Supervised Semantic Segmentation}

In this section, we discuss the algorithms for weakly-supervised semantic segmentation using image-level labels, including single- and multi-stage methods.

\textbf{Single-stage methods.} Compared with multi-stage models, single-stage methods \cite{paperRNN,paperEM,paperMIL,papertrans,paperjoint} do not have complicated training process. Due to its simplicity, these methods barely achieve sufficient segmentation accuracy. Recently, RRM \cite{paper4} and SSSS \cite{paperSSSS} tailor appropriate steps and gain accuracy similar to that of multi-stage methods.  RRM \cite{paper4} generates the initial seeds during the model training, combining the information of shallow features to optimize the segmentation network. SSSS \cite{paperSSSS} presents a segmentation-based network and a self-supervised training scheme to deal with the inherent defects of CAMs.

\textbf{Multi-stage methods.} Existing multi-stage approaches firstly generate pixel-level seeds from image-level labels by CAM (or grad-CAM), and then refine these seeds. Similarly, several approaches \cite{paper19,paper37,paper40,paper29,paper1} refine the initial cue via expanding the region of attention. SEC \cite{paper19} uses three principles, i.e., seed, expand, and constrain, to refine the initial predictions. To improve the network training, MCOF \cite{paper37} adopts a top-down and bottom-up framework to alternatively expands the labeled regions and optimizes the segmentation model. Similarly, the DSRG approach \cite{paper29} refines the initial localization maps by adopting a seeded region growing approach during the network training. The MDC method \cite{paper40} expands the seeds via utilizing multi-branches of convolutional layers with different dilation rates.

On the other hand, CAMs are mainly trained in a classification task and tend to be activated by part of the object, resulting in less satisfactory prediction.  Researchers are committed to improving the quality of the initial seeds. Adversarial erasing \cite{paper16,paper38} is a popular CAM extension method, which erases the most discriminative part of CAMs, guides the network to learn classification features from the other regions, and expands activation maps. Instead of using the erasing scheme, SEAM \cite{paper2} presents a self-supervised equivariant attention mechanism to reduce the semantic gap between the segmentation and the classification.  ScE \cite{paper5} adopts sub-category classification tasks to cluster the foreground regions. More recently, MCIS \cite{paper3} adopts semantic attention information across images to locate objects. To separate the foreground objects from the background, ICD \cite{paper6} proposes an efficient intra-class discrimination approach.

In summary, initial labels play a pivotal role on the image-level WSSS. Using this concept, we apply the concept of multi-stage methods in single-stage image-level WSSS in order to further improve the segmentation accuracy.
\subsection{Learning pixel-level affinities}
The deploy of pairwise pixel affinity has a long history in image segmentation \cite{paper2000,paper2003}, which has a variety of variants recently. For example, in fully-supervised segmentation, AAF \cite{paper44} uses adaptive affinity fields to capture and match the relations between neighboring pixels in the label space. Similarly, AffinityNet \cite{paper1} considers pixel-wise affinity to propagate local responses to the neighbouring regions, one of the popular methods for refining initial seeds. Moreover, \cite{paper12,paper13} explore cross-image relationships to collect complementary information during inference. However, errors often exist in the generated semantic labels, while the `fake' samples consequently adopt the same confidence to reinforce themselves in an affinity learning scheme, leading to over-fitting.

\subsection{Metric Learning}

Metric learning methods \cite{papernpair,papercenter,paper29} aim to optimize the transferable embeddings by learning a distance-based prediction rule over the embeddings. One advantage of using metric learning is to learn better representations of points and transfer knowledge as much as possible in the embedding space. For example, center loss \cite{papercenter} has been brought up to serve as an auxiliary loss for softmax loss to learn more discriminative features. On the other hand, triplet loss \cite{paper29} encourages embeddings of data points with the same label to get closer than those with different identities. Several variants of this work have also been proposed, such as \cite{paper20,paper23,paper27}. To reduce the time-consuming mining of hard triplets and tremendous data expansion, a novel loss named triplet-center loss \cite{paper21} is put forward. Furthermore, \cite{paper101} employs a metric learning network to transfer information from limited labeled samples to a large number of unlabeled samples in the embedding space.

\section{Proposed Approach}

Let $\mathcal{I}=\left\{\left(\boldsymbol{I}_{i}, \boldsymbol{Y}_{i}\right)\right\}_{n}$ be the set of training data, where $\boldsymbol{I}_{n}$ is the $n^{t h}$ input image, and $Y_{n} \in\{0,1\}^{C}$ is the associated ground-truth image label for $C$ semantic categories. As shown in Figure \ref{fig. geng}, data $(\boldsymbol{I}_{i},\boldsymbol{Y}_{i})$ are sampled from $\mathcal{I}$ for training the classification branch, After having fed CAM and pseudo labels to the segmentation branch,  $\boldsymbol{F}^{s} \in \mathbb{R}^{H \times W \times C_{1}}$ and $\boldsymbol{F}^{c} \in \mathbb{R}^{H \times W \times C_{2}}$ are obtained, each with the spatial dimension of $H \times W$ and different channel dimensions. In this work, we adopt the RRM framework \cite{paper4} with classification and cross entropy losses as our baseline. As mentioned before, we transplant the concept of multi-stage methods in a single-stage image-level WSSS to improve the segmentation performance. In our approach,  there are four components in the objective function: (1) Adaptive affinity loss, (2) label reassign loss, (3) standard cross entropy loss and (4) standard classification loss. The former two reflect the quality of initial labels and the relationship between image pixels.



\begin{figure*}[!t]
	\begin{center}
		\centering
		\subfigure[Image]{
			\begin{minipage}[t]{0.16\linewidth}
				\centering
				\includegraphics[scale=0.5 ]{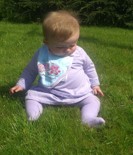}
			\end{minipage}%
		}
		\subfigure[Initial label
		]{
			\begin{minipage}[t]{0.13\linewidth}
				\centering
				\includegraphics[scale=0.5 ]{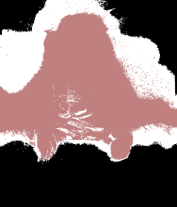}
			\end{minipage}%
		}
		\subfigure[Epoch 2]{
			\begin{minipage}[t]{0.13\linewidth}
				\centering
				\includegraphics[scale=0.5 ]{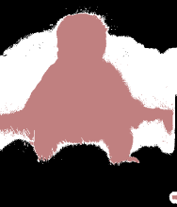}
			\end{minipage}%
		}
		\subfigure[Epoch 4]{
			\begin{minipage}[t]{0.13\linewidth}
				\centering
				\includegraphics[scale=0.5 ]{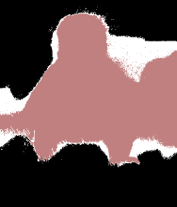}
			\end{minipage}%
		}
		\subfigure[Epoch 6]{
			\begin{minipage}[t]{0.13\linewidth}
				\centering
				\includegraphics[scale=0.5 ]{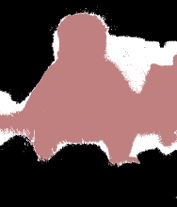}
			\end{minipage}%
		}
		\subfigure[Epoch 8]{
			\begin{minipage}[t]{0.13\linewidth}
				\centering
				\includegraphics[scale=0.5 ]{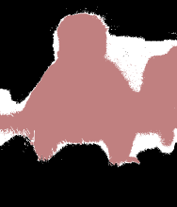}
			\end{minipage}%
		}
		\hfill
		
		\subfigure[Ground Truth]{
			\begin{minipage}[t]{0.16\linewidth}
				\centering
				\includegraphics[scale=0.5 ]{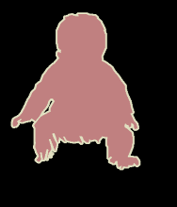}
			\end{minipage}%
		}
		\subfigure[Initial label
		]{
			\begin{minipage}[t]{0.13\linewidth}
				\centering
				\includegraphics[scale=0.5 ]{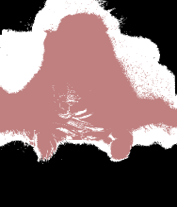}
			\end{minipage}%
		}
		\subfigure[Epoch 2]{
			\begin{minipage}[t]{0.13\linewidth}
				\centering
				\includegraphics[scale=0.5 ]{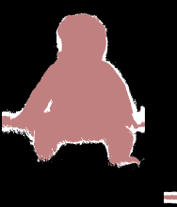}
			\end{minipage}%
		}
		\subfigure[Epoch 4]{
			\begin{minipage}[t]{0.13\linewidth}
				\centering
				\includegraphics[scale=0.5 ]{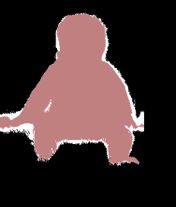}
			\end{minipage}%
		}
		\subfigure[Epoch 6]{
			\begin{minipage}[t]{0.13\linewidth}
				\centering
				\includegraphics[scale=0.5 ]{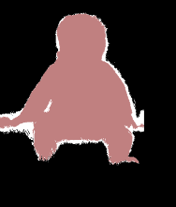}
			\end{minipage}%
		}
		\subfigure[Epoch 8]{
			\begin{minipage}[t]{0.13\linewidth}
				\centering
				\includegraphics[scale=0.5 ]{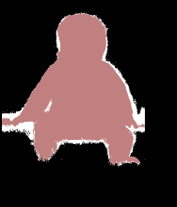}
			\end{minipage}%
		}
		
	\end{center}
	\caption{The refined results of initial labels with different epochs. The first row: outcome of the baseline. The second row: outcome of our proposed method.}
	\label{fig. gen2}
\end{figure*}

\subsection{Adaptive Affinity Loss}


To adopt the labeled samples for the generation of pseudo masks, some multi-stage approaches have used affinity learning so that semantic relationships can be transferred from the initial labels to the other regions. We adopt this concept in our single-stage image-level WSSS so as to deliver accurate semantic information.

\textbf{Standard Affinity Loss} Our implementation is derived from the idea of AAF \cite{paper44} which adopts affinity learning for network predictions instead of adding another network. Following the practice of AAF \cite{paper44}, we take advantage of the Standard Affinity (SA) loss, combined with cross entropy loss and semantic label correlation. Consequently, the SA loss encourages the network to learn inter- and inner-class pixel relationships within the labeled regions (i.e., for the pixels where initial labels and unlabeled regions are ignored).

\begin{figure}[t]
	\centering
	\includegraphics[width=8.435cm, height=3.878cm]{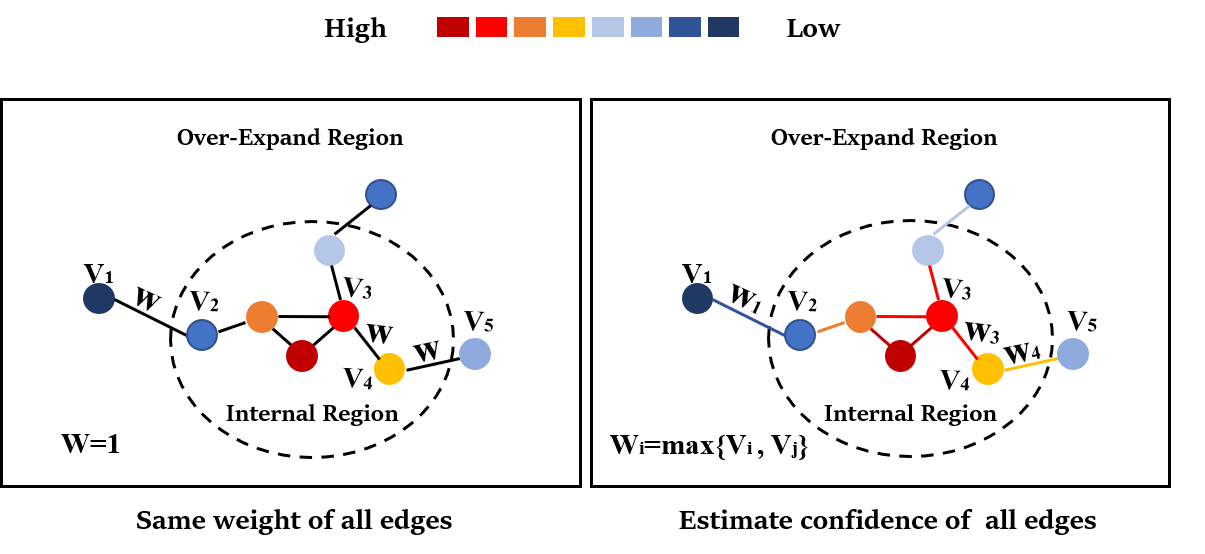}\\
	\caption{The major difference between our proposed adaptive affinity learning and standard affinity learning is that the former combines edge connectivity with the activated confidence score of the corresponding vertices.}
	\label{fig.G3}
\end{figure}
Following the setting of \cite{paper1}, depending on whether the two pixels belong to the same label or the region they appear, the set of pixel pairs $\mathcal{P} $ are divided into three subsets, where $\mathcal{P}_{\mathrm{fg}}^{+}$ is set up for the same object label in the foreground, $\mathcal{P}_{\mathrm{bg}}^{-}$ for the same label in the background, and $\mathcal{P}^{-}$ refers to different label pairs.  Moreover, similar to AAF \cite{paper44}, we combine multiple 3 * 3 kernels with varying dilation rates to learn pixel relationships over the all locations.

To obtain the SA loss, we expect to apply our loss function with the same kernel size. Consequently, the Single Kernel Standard Affinity (SKSA) loss is defined in three cases:
\begin{equation}
\begin{aligned}
\mathcal{L}_{\mathrm{fg}}^{+} &=\frac{1}{\left|\mathcal{P}_{\mathrm{fg}}^{+}\right|} \sum_{(i, j) \in \mathcal{P}_{\mathrm{fg}}^{+}} W_{i j} \\
\end{aligned}
\end{equation}
\begin{equation}
\begin{aligned}
\mathcal{L}_{\mathrm{bg}}^{+} &=\frac{1}{\left|\mathcal{P}_{\mathrm{bg}}^{+}\right|} \sum_{(i, j) \in \mathcal{P}_{\mathrm{bg}}^{+}} W_{i j}  \\
\end{aligned}
\end{equation}
\begin{equation}
\begin{aligned}
\mathcal{L}^{-} &=\frac{1}{\left|\mathcal{P}^{-}\right|} \sum_{(i, j) \in \mathcal{P}^{-}} \max \left(0, m- W_{i j}\right)
\end{aligned}
\end{equation}
where $W_{i j}$ is the Kullback-Leibler divergence between the prediction probabilities. $m$ is the margin of the separating force.
\begin{equation}
W_{i j} = D_{K L}\left(\boldsymbol{F}^{c}_{i} \| \boldsymbol{F}^{c}_{j}\right)
\end{equation}
Furthermore, the SKSA loss $\mathcal{L}_{\mathrm{SKSA}}$ can be obtained as:
\begin{equation}
\mathcal{L}_{\mathrm{SKSA}}=\mathcal{L}_{\mathrm{fg}}^{+}+\mathcal{L}_{\mathrm{bg}}^{+}+2 \mathcal{L}^{-}
\end{equation}
Finally, the SA loss is defined as:
\begin{equation}
\mathcal{L}_{\mathrm{SA}}=\sum_{k\in K} \mathcal{L}_{\mathrm{SKSA}}^{k}
\end{equation}
where the SA loss is superposed with different dilation rates $K$.

However, compared with the fully supervised semantic segmentation, initial labels out of single-stage methods may be of errors and unlabeled samples may be added to the process. In addition, the initial labels may be changed during the network training for the single-stage methods. Hence, we here propose an adaptive learning method to measure the validity of the affinity connections.

It is a common sense that the regions within the object or background incline to draw higher confidence scores in the initial predictions, while the regions near the borders receive lower scores.  As illustrated at the left hand side of Figure
\ref{fig.G3}, the standard learning methods assign the same connection weight to all the labeled samples. However, semantic confounding occurs frequently nearby the boundaries due to the misleading labels, producing negative impacts on semantic communication. The network learns the incorrect semantic relationship and keeps reinforcing it, leading to worse outcomes.

To address this issue, we propose to measure the connectivity of affinity based on the pixel activated score in order to enforce affinity learning on the region which is more reliable (i.e., internal region). As a result, we use the Adaptive Affinity (AA) loss to cooperate with pixel-wise cross entropy loss to better learn the pixel relationships.

In particular, we hypothesize that one pixel with a lower activated score is more unlikely to be true due to unpredictable semantic confounding. Given a sample $\left(V_{i}, V_{j}\right)$ with $V_{i}$ (the confidence activated score of pixel $i$) and $V_{j}$ (the confidence activated score pixel $j$), we describe the probability of connectivity using a simple yet effective maximization function, defined as:
\begin{equation}
 \Omega_{i j} = \max\left(V_{i}, V_{j}\right),
\end{equation}
where the weight of connection $\Omega_{i j}$ is defined as the higher
confidence score of the two points to penalize the connections within the boundaries.


Similar to the SKSA loss, the Single Kernel Adaptive Affinity (SKAA) loss is also defined in three cases:
\begin{equation}
\begin{aligned}
\mathcal{L}_{\mathrm{fg}}^{+} &=\frac{1}{\left|\mathcal{P}_{\mathrm{fg}}^{+}\right|} \sum_{(i, j) \in \mathcal{P}_{\mathrm{fg}}^{+}} \Omega_{i j} * W_{i j} \\
\end{aligned}
\end{equation}
\begin{equation}
\begin{aligned}
\mathcal{L}_{\mathrm{bg}}^{+} &=\frac{1}{\left|\mathcal{P}_{\mathrm{bg}}^{+}\right|} \sum_{(i, j) \in \mathcal{P}_{\mathrm{bg}}^{+}} \Omega_{i j} *  W_{i j}  \\
\end{aligned}
\end{equation}
\begin{equation}
\begin{aligned}
\mathcal{L}^{-} &=\frac{1}{\left|\mathcal{P}^{-}\right|} \sum_{(i, j) \in \mathcal{P}^{-}} \max \left(0,  (\Omega_{i j}*m- W_{i j})\right)
\end{aligned}
\end{equation}

Here, the SKAA loss $\mathcal{L}_{\mathrm{SKAA}}$ is defined as:
\begin{equation}
\mathcal{L}_{\mathrm{SKAA}}=\mathcal{L}_{\mathrm{fg}}^{+}+\mathcal{L}_{\mathrm{bg}}^{+}+2 \mathcal{L}^{-}
\end{equation}

Finally, the Adpative Affinity (AA) loss based on different dilation rates $K$ is written as:
\begin{equation}
\mathcal{L}_{\mathrm{AA}}=\sum_{k \in K} \mathcal{L}_{\mathrm{SKAA}}^{k}
\end{equation}

\subsection{Erroneous Pseudo-Label Refinement}

The  detail of the label reassign loss is illustrated in the top-right of Figure \ref{fig. geng}, which learns semantic similarity between each feature embedding and the centroid embedding of each class to mitigate over-fitting to erroneous labels. For simplicity, we take the triplet-center loss \cite{paper21} as our prototype for model learning.


\textbf{Motivation.} As reported in \cite{papernorm}, applying only few but correct labels may leads to satisfactory segmentation results. Although most methods apply high confidence thresholds to acquiring the initial predictions or updated regions in the iterative process, erroneous labels still exists somewhere. Upon this occurrence, the network suffers from over-fitting in the training process. To alleviate the above problem, we propose a novel Label Reassign (LR) loss based on metric learning.

\textbf{Centroid Calculation.}
In order to better enhance the prototype, we calculate the per-class centroids related to network predictions, making the optimization result more robust. The per-class centroid is the weighted mean of its labeled samples in the embedding space:
\begin{equation}
c_{k}=\frac{1}{\sum_{i=1}^{N} \beta_{i}} \sum_{x_{i} \in l_{k}} \beta_{i} * \boldsymbol{F}^{s}_{x_{i}},
\end{equation}
where $\beta_{i}$ is the confidence value of the labeled pixel
embedding $x_{i}$, which belongs to the set of feature embedding $l_{k}$, and $\boldsymbol{F}^{s}_{x_{i}}$ is the embedding vector. Here, the confidence score can be obtained through calculating the network prediction probability. Then, the label of each pixel can be assigned to the class with the highest similarity value.

\textbf{Label Reassign Loss.}
To address the problem of class imbalance, we split the embedding points into two parts, $E_{b g}$ and $E_{f g}$ for the background and the foreground, respectively. Then, the metric-based loss is computed as follows:
\begin{equation}
\mathcal{L}_{-}=\frac{1}{\left|E_{\mathrm{bg}}\right|}\sum_{x_{i} \in E_{\mathrm{bg}}}  \alpha_{i}* \sum_{j=1}^{N}\max \left(0, n+D_{x_{i} c_{j}}-D_{x_{i} c_{i}}\right)
\end{equation}
\begin{equation}
\mathcal{L}_{+}=\frac{1}{\left|E_{\mathrm{fg}}\right|} \sum_{x_{i} \in E_{\mathrm{fg}}} \alpha_{i}*\sum_{j=1}^{N}\max \left(0, n+D_{x_{i} c_{j}}-D_{x_{i} c_{i}}\right)
\end{equation}
Where  $N$ is the class number (including the background) in each image, which is greater than or equal to 2, $D(*)$ represents the cosine similarity and $\alpha_{i}$ is the modulation parameter for feature embedding $i$.

\begin{equation}
\alpha_{i}=\left(1 - \frac{D_{x_{i} c_{i}}-D_{x_{i} c_{j}}}{D_{x_{i} c_{i}}+D_{x_{i} c_{j}}}\right)^{\gamma},
\end{equation}
where $D_{x_{i} c_{i}}$ is the distance between the reassigned pixel embedding and its nearest centroid $c_{i}$, while $c_{j}$ is the second nearest centroid to $x_{i}$.
$\gamma$ is a concentration parameter. When $\gamma$ = 0, the LR loss is identical to the prototype, and as $\gamma$ increases, the LR loss focuses on the samples with a closer distance between $D_{x_{i} c_{i}}$ and $D_{x_{i} c_{j}}$, more likely to be affected by the erroneous labels.


Finally, the metric-based LR loss is defined as:
\begin{equation}
\mathcal{L}_{\mathrm{LR}}=\mathcal{L}_{\mathrm{-}}+\mathcal{L}_{\mathrm{+}}
\end{equation}

\section{Experimental work}


\subsection{Implementation Details}

\textbf{PASCAL VOC 2012.} PASCAL VOC 2012 segmentation dataset is of 21 class annotations, i.e., 20 foreground objects and the background. The official dataset has 1464 images for training, 1449 for validation and 1456 for testing. Following the common experimental protocol for semantic segmentation, we take additional annotations from SBD \cite{paper14} to build an augmented training set with 10582 images. Noting that only image-level classification labels are available during the network training. Mean intersection over union (mIoU) is considered as the evaluation criterion.

\textbf{Initial mask generation.} Similar to \cite{paperSSSS}, we remove the masks for the classes with classification confidence $<$ 0.1. Following the common practice \cite{paper1,paper4}, we generate the initial masks after having applied dCRF \cite{paperdcrf} during the training.

\textbf{Backbone Framework $\&$ Network.} In order to validate the network, we adopt the proposed losses on RRM \cite{paper4} with default parameters as our backbone framework due to its superior performance on single-stage WSSS and encouraging training efficiency. Identical to \cite{paper4}, we also use a WideResNet38 backbone network \cite{paperresnet38} provided by \cite{paper1}.

\textbf{Data augmentation.}  Similar to RRM \cite{paper4}, we use random rescaling (falling in the range of (0.7, 1.3) w.r.t. the original image area) and horizontal flipping. Finally, the images are cropped to the size of 321 * 321.

\textbf{Training Time $\&$ Memory.} Our method requires a slight increase (less than 4$\%$) in training time compared to RRM \cite{paper4} (a single-stage technique). Meanwhile, we reduce a large quantity of training time in comparison with all the multi-stage methods, especially NSROM \cite{yao2021non}, due to their complex processing operations. The system is implemented in PyTorch and trained on four NVIDIA 1080ti GPUs with 11 GB memory. The testing is conducted on the same machine with one GPU.

\textbf{Training.} The final training objective is to combine all the above mentioned objectives, i.e.,  $L_{total}=\mathcal{L}_{\mathrm{CLS}}+\mathcal{L}_{\mathrm{CE}}+\lambda_{1} \mathcal{L}_{\mathrm{AA}}+\lambda_{2} \mathcal{L}_{\mathrm{LR}}$. We
empirically set $\lambda_{1}=0.1, \lambda_{2}=0.1 .$ For the adaptive
affinity loss, margin $m$ is set to 3.0 with the 4-8-16-24 kernel size. For the label reassign loss, threshold $n$ is set to 1 and used in the last two epochs to prevent the occurrence of over-fitting. For the feature dimension, we use $C_{1}$ = 512,  $C_{2}$ = 21. We train our model with the initial learning rate of 1e-4 and the weight decay of 5e-4.

\begin{figure}[t]
	\centering
	\includegraphics[width=8.1405cm, height=5.211cm]{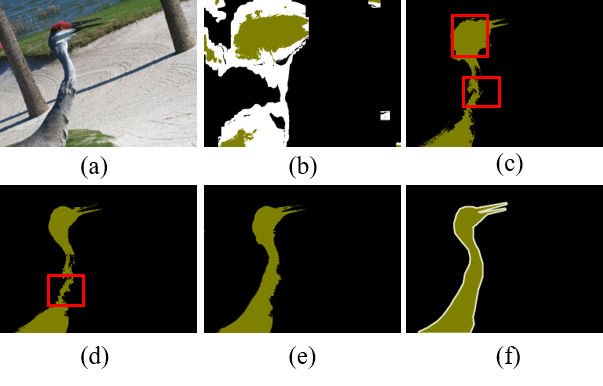}\\
	\caption{Visualization results of adaptive pairwise affinity learning. (a) Raw images. (b) Initial labels. (c) Baseline with only CE and CLS losses. (d) Baseline + SA loss (w/o Adaptive). (e) Baseline + AA loss. (f) Ground truth. Red rectangles highlight the results of incorrect semantic propagation.}
	\label{fig.FPX}
\end{figure}

\begin{figure}[t]
	\centering
	\includegraphics[width=8.0629164cm, height=7.17114024cm]{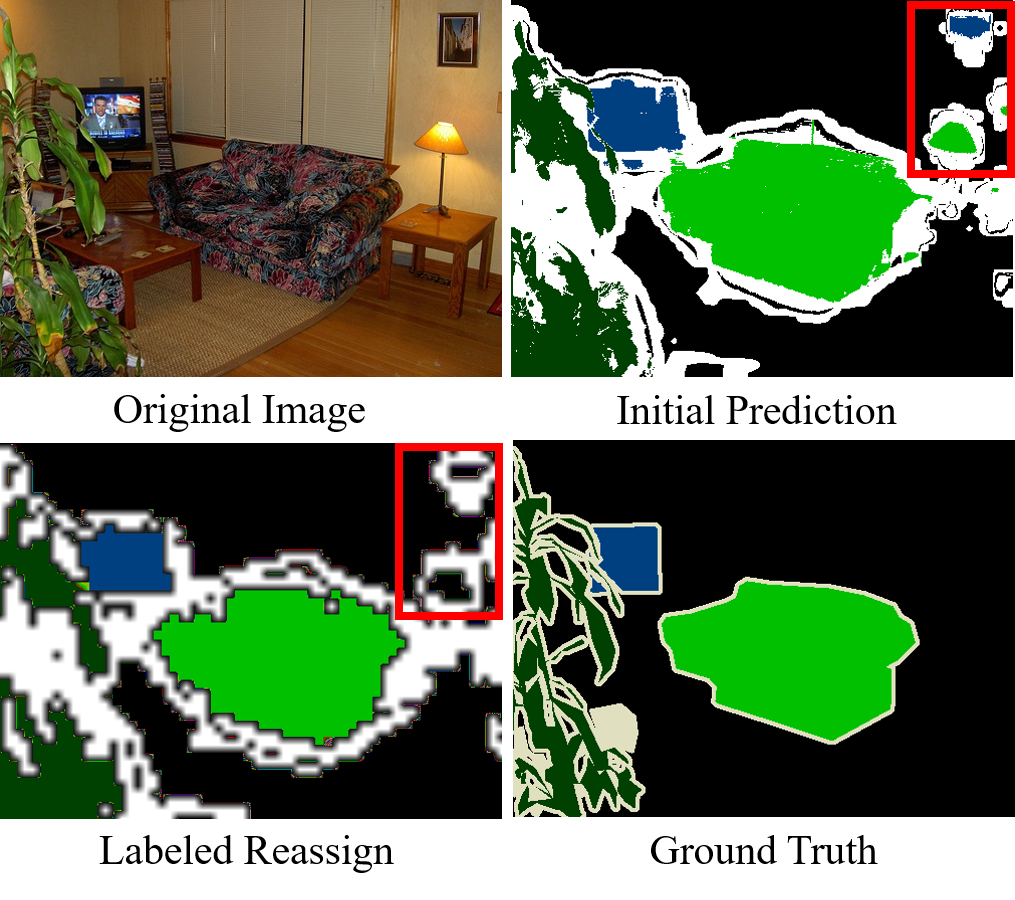}\\
	\caption{Visualization results of the label reassign (upsampling from feature map). Red rectangles highlight the improved regions predicted by label reassign which avoids over-fitting on noisy labels.}
	\label{fig.G2}
\end{figure}

\begin{figure*}[t]
	\centering
	\includegraphics[width=17.3536cm, height=12.2844cm]{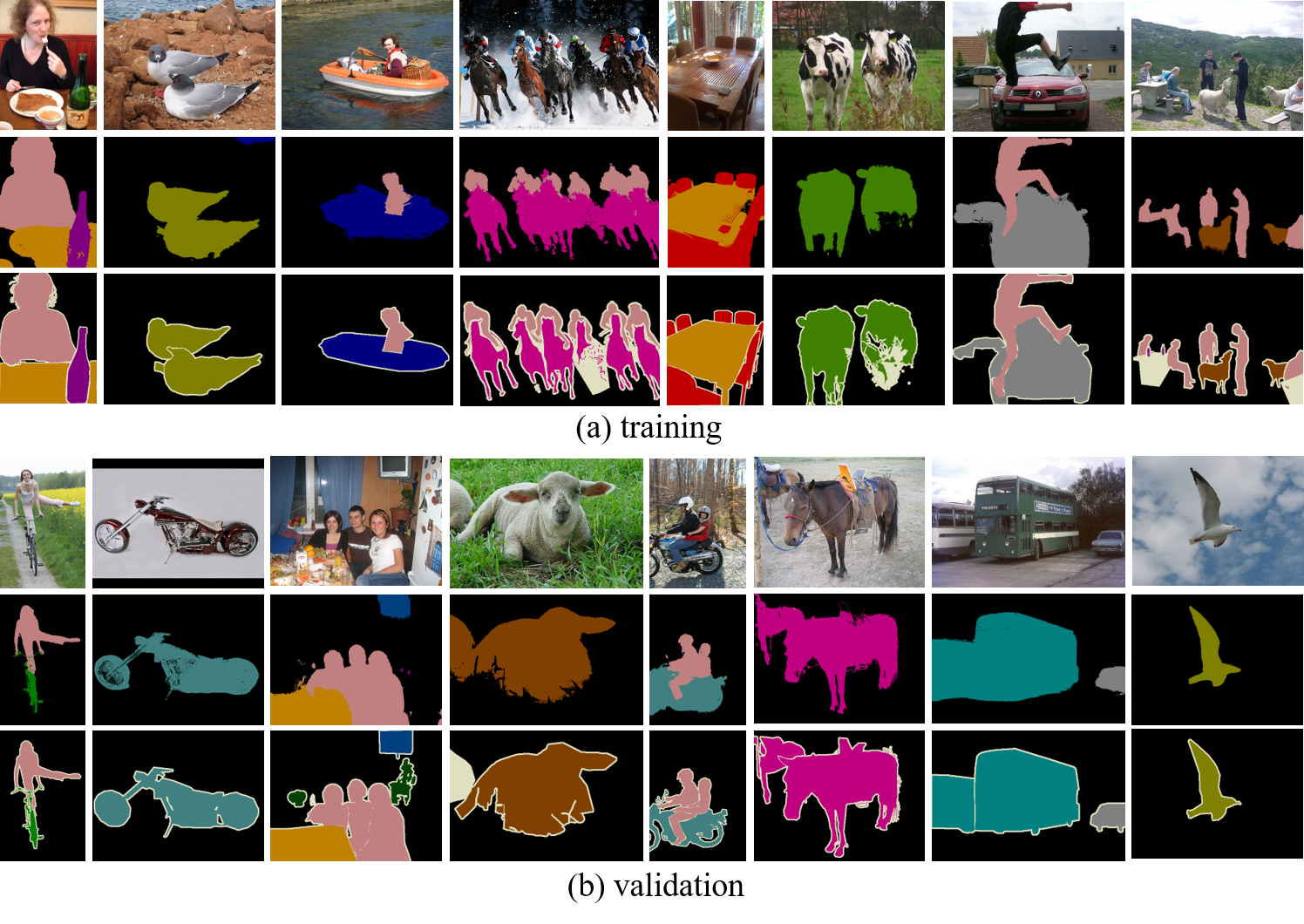}\\
	\caption{ Qualitative results of our single-stage method on the PASCAL VOC 2012 dataset. From top to bottom of (a) training and (b) validation: input images, our segmentation results, and ground-truths.}
	\label{tab.knight6}
\end{figure*}

\begin{figure}[t]
	\centering
	\includegraphics[width=8.0975cm, height=5.7038cm]{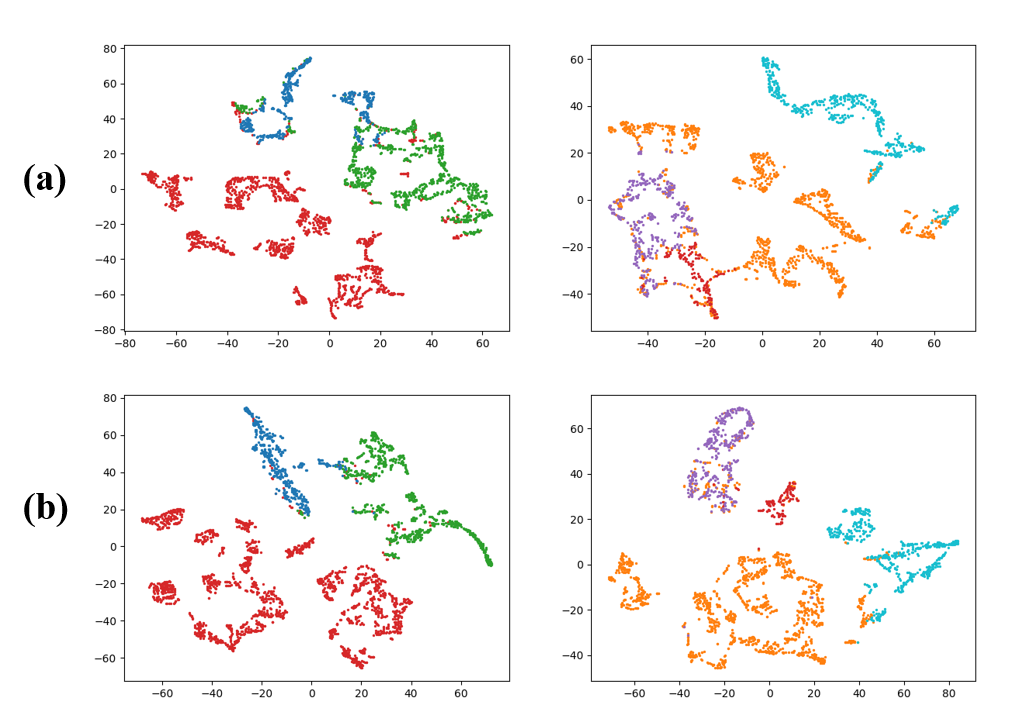}\\
	\caption{T-SNE visualization of point embeddings in the 2D space. \textbf{(a)} Feature embedding w/o label reassign \textbf{(b)} Feature embedding with label reassign.}
	\label{fig.FPX2}
\end{figure}


\renewcommand{\arraystretch}{1.3}
\begin{table}[t]
	\caption{\textbf{Ablation study on Pascal VOC dataset}. We study the impact of individual losses. Additionally, same as \cite{paperSSSS}, we adopt ground-truth image-level labels to remove masks of any false positive class predicted by our model with $\checkmark$ in the ``rfpc'' column.}	
	\begin{center}
		\begin{tabular}{ccccccc}
			\toprule[1pt]
			CLS & CE & AA & LR & rfpc & \textbf{IoU} (train) & \textbf{IoU} (val)  \\ \bottomrule[1pt]
			\checkmark   & \checkmark  &     &  &  & 63.7 & 60.0 \\ \hline
			\checkmark   & \checkmark  & \checkmark   &  &  & 66.7 & 63.2 \\ \hline
			\checkmark   & \checkmark  & \checkmark   & \checkmark & & 67.4 & 63.9 \\ \hline
			\checkmark   & \checkmark  & \checkmark   & \checkmark & \checkmark  & \textbf{68.2} & \textbf{65.8} \\ \bottomrule[1pt]
		\end{tabular}
	\end{center}

	\label{tab.knight1}
\end{table}

\renewcommand{\arraystretch}{1.3}
\begin{table}[t]
\caption{\textbf{IoU on Pascal VOC dataset}. w.r.t. the adaptive pairwise affinity loss. We compare the quality of SA (w/o adaptive) and the AA loss with different kernel sizes.}
	\label{tab.jkl}
	\begin{center}
		\scalebox{0.95}{
			\begin{tabular}{c|lllll|ll}
				\toprule[1pt]
				Size           & \multicolumn{1}{c}{4} & \multicolumn{1}{c}{8} & 12 & 24 & 36 & \textbf{IoU} (train)               & \textbf{IoU} (val)                 \\ \hline
				& \checkmark                     & \checkmark                     & \checkmark  & \checkmark  & \checkmark  & \multicolumn{1}{c}{65.3} & \multicolumn{1}{c}{61.6} \\
				SA & \checkmark                     & \checkmark                    & \checkmark  & \checkmark  &    &  \multicolumn{1}{c}{66.0}                        &     \multicolumn{1}{c}{62.3}                     \\
				& \checkmark                     & \checkmark                    & \checkmark  &    &    & \multicolumn{1}{c}{65.9}                         & \multicolumn{1}{c}{61.9}                         \\ \toprule[1pt] \toprule[1pt]
				& \checkmark                     & \checkmark                    & \checkmark  & \checkmark  & \checkmark  & \multicolumn{1}{c}{65.6} & \multicolumn{1}{c}{62.0} \\
				AA              & \checkmark                     & \checkmark                     & \checkmark  & \checkmark  &    & \multicolumn{1}{c}{\textbf{66.7}} & \multicolumn{1}{c}{\textbf{63.2}} \\
				& \checkmark                     & \checkmark                    & \checkmark  &    &    &  \multicolumn{1}{c}{66.2}                        &\multicolumn{1}{c}{62.4}                          \\ \bottomrule[1pt]
				
		\end{tabular}}
	\end{center}

\end{table}
\renewcommand{\arraystretch}{1.3}
\begin{table}[t]
\caption{\textbf{IoU on Pascal VOC dataset}. w.r.t. the adaptive pairwise affinity loss. We compare different modeling functions using the kernel size 4-8-12-24. $\bm{plus\left(V_{i}, V_{j}\right)}$ means $\bm{\left(V_{i}+ V_{j}\right)/2}$.}
	\begin{center}
		\begin{tabular}{c|ccc}
			\toprule[1pt]
			MF & $\bm{max\left(V_{i}, V_{j}\right)$} & $min\left(V_{i}, V_{j}\right)$ &$plus\left(V_{i}, V_{j}\right)$ \\ \bottomrule[1pt]
			\textbf{IoU} (train) & \textbf{66.7}       & 66.1   &  66.4    \\ \hline
			\textbf{IoU} (val) &  \textbf{63.2}         & 62.5    & 62.7    \\ \bottomrule[1pt]
			
		\end{tabular}
	\end{center}

	\label{tab.doinb}
\end{table}

\renewcommand{\arraystretch}{1.3}
\begin{table}[t]
\caption{\textbf{IoU on Pascal VOC dataset}. w.r.t. the label reassign loss. We compare the quality of using different concentration parameters.}
	\label{tab.doinb2}
	\begin{center}
		\begin{tabular}{c|ccccc}
			\toprule[1pt]
			$\gamma$ &0 & 0.5 & 1 & 2 & 5\\ \bottomrule[1pt]
			\textbf{IoU} (train)&66.9 & 67.1         & 67.2   & \textbf{67.4} & 67.0   \\ \hline
			\textbf{IoU} (val) &63.5 &  63.6         & 63.7    & \textbf{63.9} & 63.4  \\ \bottomrule[1pt]
			
		\end{tabular}
	\end{center}

\end{table}



\renewcommand{\arraystretch}{1.5}
\begin{table*}[t]
\caption{Performance on the PASCAL VOC 2012 validation set by several single-stage methods.}
	
	\begin{center}
		\resizebox{\textwidth}{!}{
			
			\begin{tabular}{clllllllllllllllllllll|l}
				\hline
				single stage Method & bkg & plane & bike & bird & boat & bottle & bus & car & cat & chair & cow & table & dog & horse & mbk & person & plant & sheep & sofa & train & tv & mIoU \\ \hline
				RRM \cite{paper4}	&  87.9   &   75.9    &  31.7    &  78.3    &  54.6      &  62.2   &  80.5   &  73.7   &   71.2    &  30.5   &  67.4     &  40.9   &  71.8     &  66.2   &   70.3     &   72.6    &   49.0    &   70.7   &  38.4     &  62.7  &  58.4 & 62.6    \\ \hline
				SSSS \cite{paperSSSS}	&  88.7   &    70.4   &  35.1    &  75.7    &   51.9     &  65.8   &  71.9   &  64.2   &  81.1     & 30.8    & 73.3      &  28.1   &   81.6    &  69.1   &   62.6     &  74.8     &  48.6     & 71.0     &  40.1     & 68.5   & 64.3  & 62.7  \\ \hline
				Ours	&   88.4  &  \textbf{76.3}     &  33.8    & \textbf{79.9}   &   34.2     &  \textbf{68.2}   & 75.8    & \textbf{74.8}    &   \textbf{82.0}   &  \textbf{31.8}   &  68.7     &  \textbf{47.4}   & 79.1      &  68.5   &   \textbf{71.4}     &   \textbf{80.0}    &   \textbf{50.3}    &   \textbf{76.5}   &  \textbf{43.0}     &  55.5  &  58.5 & \textbf{63.9}    \\ \hline
		\end{tabular}}
		
	\end{center}
    \label{tab.knight4}
\end{table*}
\subsection{Ablation Study and Analysis}

We analyze the importance of the proposed losses, as shown in Table \ref{tab.knight1}. By adding the proposed losses, the baseline can be improved significantly. Since the CNNs are not responsible for the delivery of correct semantic information, the adaptive affinity loss can lead us to better understanding this problem (from 63.7$\%$ to 66.7$\%$ IoU in the training set and 60.0 $\%$ to 63.2$\%$ IoU in validation set). Moreover, the label reassign loss mitigates over-fitting on the erroneous labels, further improving the network performance (from 66.7$\%$ to 67.4$\%$ IoU in training set and 63.2 $\%$ to 63.9$\%$ IoU in validation set). Finally, following the practice of \cite{paperSSSS}, we remove any false positive class prediction to achieve the best results. As shown in Table \ref{tab.jkl}, compared with the standard affinity loss, by adding adaptive connectivity, our approach significantly improves the IoU  by 0.7 $\%$ in the training set and 0.9 $\%$ in the validation set.


\textbf{Labeled Regions Refinement.}
As shown in Figure \ref{fig. gen2}, the baseline has little improvement on initial labels and over-fitting at the end of the network training, leading to the degradation of segmentation performance. On the other hand, our proposed method maintains correct semantic information, assigns correct labels in the unlabeled regions, and further improves the segmentation accuracy.

\textbf{Adaptive Affinity Loss.}
In order to show the benefit of the proposed AA loss, we illustrate a set of comparison plots. As shown in Figure \ref{fig.FPX} (b), the initial labels have a number of erroneous prediction points. First, in Figure \ref{fig.FPX} (c),  if we merely fit these points, the erroneous semantics will be passed to the unlabeled regions. This makes the results worse. Next, in Figure \ref{fig.FPX} (d), with  pairwise affinity learning, the proposed framework learns the additional semantic information to set up reliable communication with the unlabeled regions, but over-fits to the wrong labeled regions as well. Finally, in Figure \ref{fig.FPX} (e), with the introduction of weights, the network learns the correct semantic information while mitigating over-fitting, and generates the results close to the ground truth.

\textbf{Modeling Function.}
To better understand the contribution of the adaptive affinity loss, Table \ref{tab.doinb} shows some common modeling functions related to confidence activated scores. For function $min\left(V_{i}, V_{j}\right)$, it penalizes the connections with lower confidence scores, avoiding most error samples. However, it also reduces the connections within the internal regions, leading to a marginal increase in the system performance. Next, for function $plus\left(V_{i}, V_{j}\right)$, it shows more smoothly changed connection scores in different regions, resulting in better performance, while blurring some of the objects unexpectedly. Moreover, function $max\left(V_{i}, V_{j}\right)$ weakens the connections within the boundaries where the confidence scores of the two points are lower than the average, in which most semantic confounding occurs frequently whilst keeping clear semantic properties in the images. The significantly improved performance shows that adaptive affinity learning plays a positive role on WSSS.

\textbf{Concentration Parameter.}
The results of using the LR loss with different concentration parameters are shown in Table \ref{tab.doinb2}. The LR loss introduces one new hyperparameter $\gamma$, the concentration parameter, that controls the strength of the modulation factor. When  $\gamma$ = 0, our loss is identical to the triplet-center loss. As $\gamma$ increases, the LR loss focuses on more vulnerable samples, which are more likely to be overfitted. With $\gamma$ = 2, the LR loss improves the IoU by 0.7 $\%$ in both training and val sets.

\textbf{Label Reassign.}
In order to understand how our learning mechanism improves the identification capacity of the network, we visualize the corresponding predictions after label reassignment (i.e., reassign the labeled pixels in the embedding space). As mentioned before, the network filters out some undesired noisy points, which helps alleviating over-fitting for the `fake' labeled samples. Figure \ref{fig.G2} shows our results, compared to the initial prediction, the label reassign produces  better segmentation results. Moreover, we project the feature embedding via T-SNE \cite{paperTSNE}. The projected point embeddings are shown in Figure \ref{fig.FPX2}. Without the label reassign loss, semantic confounder exists in certain classes, demonstrating the effectiveness of our proposed loss.

\subsection{Semantic Segmentation Performance}


The proposed method performs better than the standard SSSS \cite{paperSSSS} and the RRM \cite{paper4} on the validation set. For the test set, we achieve system performance similar to that of some multi-stage methods.

In Table \ref{tab.knight4}, we show detailed results for each category on the validation set. In Figure \ref{tab.knight6}, we present some examples of the final semantic segmentation results, where our results are close to the ground truthed segmentation.

In Table \ref{tab.knight8}, we provide performance comparisons with the other  image-level WSSS methods. Our method outperforms the other state of the art single-stage WSSS methods with the mIoU scores of 63.9 and 64.8 on the PASCAL VOC 2012 validation and test sets, respectively, and achieves similar performance to the multi-stage WSSS methods.

\section{Conclusions}

In this work, we have presented two novel losses to improve the performance of a single-stage WSSS. On the one hand, we proposed the Adaptive Affinity loss to model the relationship between image pixels so that the network can carry out  accurate semantic propagation. On the other hand, we proposed the Label Reassign loss to identify the semantic information of image pixels in pseudo labels to ease the fitting of false labels. Our method produced better segmentation results than several existing single-stage methods and a range of multi-stage methods.

\renewcommand{\arraystretch}{1.3}
\begin{table}[t]
\caption{\textbf{Mean IoU on the Pascal VOC validation and test sets}, \textit{I} indicates only image-level supervision with additional data \textit{A} and saliency detection \textit{S}.}
	\begin{center}
		\scalebox{0.92}{
			\begin{tabular}{ccccc}
				\toprule[1pt]
				Method                                    & \multicolumn{1}{l}{Backbone} & \multicolumn{1}{l}{Superv.} & \multicolumn{1}{l}{val}               & \multicolumn{1}{l}{test} \\ \hline
				\multicolumn{1}{l}{\textit{Multi-stage}}  & \multicolumn{1}{l}{}         & \multicolumn{1}{l}{}        & \multicolumn{1}{l}{}                  & \multicolumn{1}{l}{}     \\ \hline
				SEC \cite{paper19}                                       & VGG16                        & \textit{I}                  & 50.7                                  & 51.1                     \\
				STC \cite{paperstc}                                       & VGG16                        & \textit{I+S}                & 49.8                                  & 51.2                     \\
				AffinityNet \cite{paper1}                               & WideResNet38                     & \textit{I}                  & 61.7                                  & 63.7                     \\
				MCOF \cite{paper37}                                      & ResNet101                    & \textit{I+S}                & 60.3                                  & 61.2                     \\
				DSRG \cite{paper29}                                      & ResNet101                    & \textit{I+S}                & 61.4                                  & 63.2                     \\
				CIAN \cite{paper12}                                      & ResNet101                    & \textit{I}                  & 64.1                                  & 64.7                     \\
				SeeNet \cite{paper16}                                    & ResNet101                    & \textit{I+S}                & 63.1                                  & 62.8                     \\
				FickleNet \cite{paperfickle}                                 & ResNet101                    & \textit{S}                  & 64.9                                  & 65.3                     \\
				OAA \cite{paperOAA}                                       & ResNet101                    & \textit{I}                  & 63.9                                  & 65.6                     \\
				ScE \cite{paper5}                                       & ResNet101                    & \textit{I}                  & 66.1                                  & 65.9                     \\
				SEAM \cite{paper2}                                     & ResNet101                    & \textit{I}                  & 64.5                                  & 65.7                     \\
				MCIS \cite{paper3}                                      & ResNet101                    & \textit{I}               &   66.2                                  & 66.9                          \\
				ICD \cite{paper6}                                       & ResNet101                    & \textit{I}                  & 64.1                                  & 64.3                       \\
				CONTA \cite{zhang2020causal}                                       & ResNet101                    & \textit{I}                  & 66.1                                  & 66.7                       \\
				GWSM \cite{li2020group}                                       & ResNet101                    & \textit{I+S}                  & 68.2                                  & 68.5                       \\
				NSROM \cite{yao2021non}                                       & ResNet101                    & \textit{I+S}                  & 68.3                                  & 68.5                       \\ \hline
				\multicolumn{1}{l}{\textit{Single-stage}} & \multicolumn{1}{l}{}         & \multicolumn{1}{l}{}        & \multicolumn{1}{l}{}                  & \multicolumn{1}{l}{}     \\ \hline
				EM \cite{paperEM}                  & VGG16    & \textit{I}                  & 38.2 & 39.6 \\
				MIL-LSE \cite{paperMIL}             & overfeat\cite{paperoverfeat}    & \textit{I}                  & 42.0 & 40.6 \\
				CRF-RNN \cite{paperRNN}             & VGG16    & \textit{I}                  & 52.8 & 53.7 \\ \hline
				TransferNet \cite{papertrans}             & VGG16    & \textit{A}                  & 52.1 & 51.2 \\
				WebCrawl \cite{paperweb}             & VGG16    & \textit{A}                  & 58.1 & 58.7 \\ \hline						
				RRM \cite{paper4}                 & WideResNet38 & \textit{I}                  & 62.6 & 62.9 \\
				SSSS \cite{paperSSSS}                & WideResNet38 & \textit{I}                  & 62.7 & 64.3 \\ \hline
				Ours                                & WideResNet38 & \textit{I}                  & \textbf{63.9} &   \textbf{64.8}   \\
				\bottomrule[1pt]
				
		\end{tabular}}
	\end{center}

	\label{tab.knight8}	
\end{table}

\bibliographystyle{ACM-Reference-Format}
\bibliography{ref}

\end{document}